\def\year{2019}\relax
\documentclass[letterpaper]{article} 
\usepackage{aaai19}  
\usepackage{times}  
\usepackage{helvet}  
\usepackage{courier}  
\usepackage{url}  
\usepackage{graphicx}  
\usepackage{booktabs}       
\usepackage{amsfonts}       
\begin{document}
\title{Near-lossless Binarization of Word Embeddings}
\author{
  Julien Tissier \and Christophe Gravier \and Amaury Habrard\\
  Univ. Lyon, UJM Saint-Etienne\\
  CNRS, Lab Hubert Curien UMR 5516\\
  42023, Saint-Etienne, France\\
  \texttt{firstname.lastname@univ-st-etienne.fr}
}

\maketitle

\begin{abstract}
Word embeddings are commonly used as a starting point in many NLP models to
  achieve state-of-the-art performances. However, with a large vocabulary and
  many dimensions, these floating-point representations are expensive both in
  terms of memory and calculations which makes them unsuitable for use on
  low-resource devices.
The method proposed in this paper transforms real-valued embeddings into binary
  embeddings while preserving semantic information, requiring only 128 or 256
  bits for each vector. This leads to a small memory footprint and fast vector
  operations. The model is based on an autoencoder architecture, which also
  allows to reconstruct original vectors from the binary ones.
Experimental results on semantic similarity, text classification and sentiment
  analysis tasks show that the binarization of word embeddings only leads to a
  loss of $\sim$2\% in accuracy while vector size is reduced by 97\%.
  Furthermore, a top-k benchmark demonstrates that using these binary vectors is
  30 times faster than using real-valued vectors.
\end{abstract}

\section{Introduction}
  \label{intro}
  Word embeddings models play a central role in many NLP applications like
  document classification \cite{joulin2016bag,conneau2016very} or sentiment
  analysis \cite{socher2013recursive,qian2016linguistically}. The real-valued
  vector representation associated to each word of a vocabulary $\mathcal{V}$
  reflects its semantic and syntactic information extracted from the language
  \cite{bengio2003neural}.  They are usually created from a large corpus by
  moving closer the vectors of words co-occurring together
  \cite{mikolov2013distributed} or factorizing the matrix containing
  co-occurrences statistics \cite{pennington2014glove}, and commonly require
  several gigabytes of memory space. For example, with a vocabulary of 2 million
  words and 300-dimensional vectors, storing the word embedding require 2.4
  gigabytes (with real values encoded with $32$-bit \verb+float+).

  \subsubsection{Motivation} Storing and running such models on embedded devices
  like cell phones is not feasible in practice, due to their limited memory and
  low computing power for floating-point arithmetic. In general, NLP
  applications running on smartphones send data to large computing servers that
  perform the calculations and return the results. The method described in this
  paper aims to reduce the size of the word embeddings models so that the
  computations can be done locally on the embedded device. The benefits are
  twofold:
  \begin{enumerate}
    \item the embedded device can run NLP applications offline;
    \item privacy is preserved since no user data is send to servers.
  \end{enumerate}

  A solution to reduce the size is to \emph{binarize} the model; either the
  learning parameters \cite{hubara2016binarized} or the vector representations
  \cite{joulin2016bag}. This paper stands in the second category. Associating
  binary codes to words allows one to speed up calculations as the vector
  operations can be done with bitwise operators instead of floating-point
  arithmetic, \textit{e.g.} computing the distance between two binary vectors
  only requires a~\verb+XOR+ and a~\verb+popcount()+
  \footnote{\texttt{popcount(n)} returns the number of bits set to $1$ in
  \texttt{n}.} operations (which are both performed with a single CPU cycle due
  to hardware optimizations in modern processors) while it requires
  $\mathcal{O}(n)$ floating-point multiplications and additions for
  $n$-dimensional real-valued vectors.  To take advantage of fast CPU optimized
  bitwise operations, the size of binary vectors \emph{has to be in adequacy
  with register sizes} (64, 128 or 256 bits).  When this criteria is met, the
  computations are much faster \cite{norouzi2012fast,subercaze2015metric}.
  Nevertheless, mapping words to binary codes is not enough as the vectors are
  then used in NLP applications.  They also need to encode semantic and
  syntactic information; the objective then becomes to find binary vectors that
  preserve the language aspects and are small enough to fit in CPU registers.

  \subsubsection{Contributions}
  This paper solves the problem of producing binary word vectors from
  real-valued embeddings while preserving semantic information. Our model is
  based on an autoencoder architecture which transforms any real-valued vectors
  to binary vectors of any size (more specifically 64, 128 or 256 bits) and
  allows one to reconstruct original vectors from the binary ones. Our binary
  vectors use \textbf{37.5 times less memory} and perform a top-k query
  \textbf{30 times faster} than real-valued vectors, with only an accuracy loss
  of $\sim$2\% on several NLP tasks. Entire source code to generate and evaluate
  binary vectors is available online
  \footnote{\scriptsize\url{https://github.com/tca19/near-lossless-binarization}}.

\section{Related work}
  \subsubsection{Dimensionality reduction} Classic dimensionality reduction
  techniques \cite{raunak2017reduction} can be used to produce smaller
  embeddings. Although the number of dimensions can be halved, the produced
  vectors still require floating-point operations when they are used. Another
  method \cite{Ling2016Word} limits the precision of each dimension to $4$ or
  $8$ bits, but the size of the vectors is not aligned with CPU register
  sizes. Other approaches \cite{shu2018compressing,chen2018kway} only store a
  small set of basis vectors which are then used as linear combinations to
  produce a vector for each word. The memory size is largely reduced
  ($\sim$98\%) because they do not store the whole embeddings matrix, but the
  vectors are still real-valued and do not benefit from the aforementionned
  calculation gains.

  \subsubsection{Binary word vectors}
  A naive approach to produce binary word embeddings is to map each value $x$ of
  a pre-trained real-valued embedding to $0$ if $x < 0$ and $1$ otherwise. This
  method does not require any training step but suffers from an important
  drawback: the binary vectors \emph{have the same dimension} as the original
  ones. To produce vectors of $64$, $128$ or $256$ bits, one has to retrain
  real-valued embedding with these dimensions, and then apply the naive
  binarization. The results of subsection \ref{ssection:naive} show that it
  is slower and less efficient than the proposed method that directly transforms
  vectors with the appropriate binary size.

  Semantic hashing is one way to find compact binary codes in order to well
  approximate nearest neighbor search in the original space. NASH
  \cite{shen2018NASH} finds a binary code to each document for fast information
  retrieval, using an end-to-end neural architecture. Locality sensitive hashing
  \cite{charikar2002similarity} uses random projections to produce binary codes
  that approximates the cosine similarity of the corresponding orginal vectors.
  However, these methods generally fail to fully preserve semantic similarities
  \cite{Xu2015convolutional}.

  Faruqui et al. \shortcite{faruqui2015sparse} propose to binarize real-valued
  vectors by first increasing the vector size to create sparse vectors, and then
  applying the naive binarization function. Although this method preserves the
  semantic similarities, the produced vectors are not really small and does not
  fit in CPU registers. Fasttext.zip \cite{joulin2016fasttext} binarizes word
  vectors by clustering them and concatenating the binary codes of the $k$
  closest centroids for each word. The binary vectors are then used in a
  document classification task but this method does not produce generic binary
  vectors that could be used in other tasks as the vocabulary size is
  drastically reduced to around $1000$ words.

  Although binary vectors can speed up vector computations, some NLP
  applications only work with real-valued vectors \cite{ma2016sequence}; being
  able to reconstruct real-valued vectors from the binary ones is therefore
  required for such applications. As a consequence, one can store the binary
  embeddings and compute on the fly real-valued vectors only for the words it
  needs instead of storing the real-valued vectors of all words, reducing the
  required memory space (\textit{e.g.} for 1M words and $128$-bit vectors,
  storing binary codes and the reconstruction matrix requires 16.1MB vs. 1.2GB
  for real-valued vectors).

\section{Binarizing word vectors with autoencoder}
  \label{binauto}

\subsection{Autoencoder architecture}
\begin{figure}[h]
    \begin{center}
      \centerline{\includegraphics[width=0.7\linewidth]{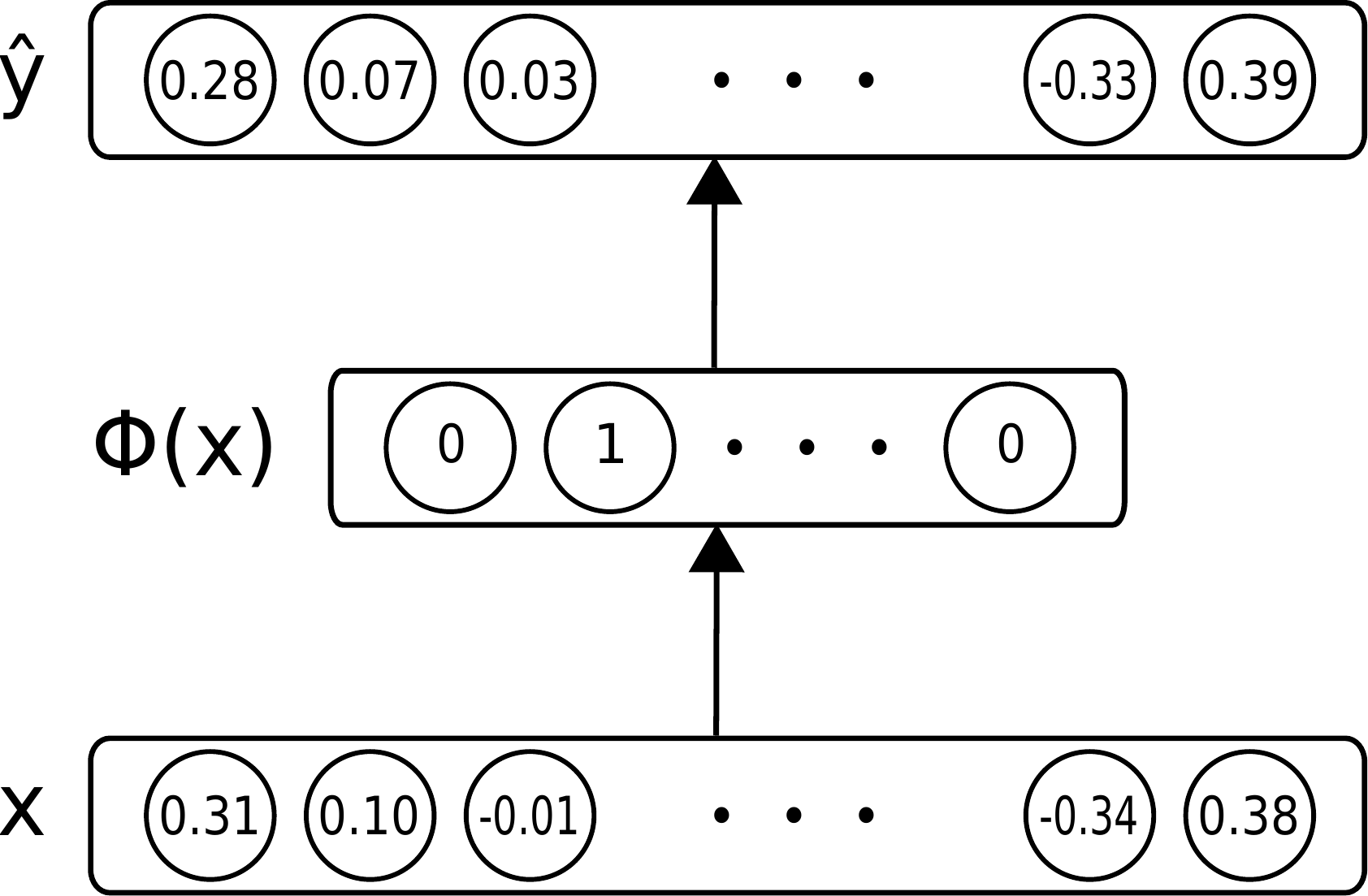}}
      \caption{Autoencoder architecture used to binarize
      word vectors and reconstruct a real-valued representation. The model can
      learn any vector size for $\Phi(x)$.}
      \label{architecture}
    \end{center}
  \end{figure}

  Let $\mathcal{V}$ be a vocabulary of words, and $X \in
  \mathbb{R}^{|\mathcal{V}| \times m}$ a $(|\mathcal{V}| \times m)$ matrix whose
  rows are $m$-dimensional real-valued vectors representing the embedding of
  each word. Our main objective is to transform each row $x_i$ into a binary
  representation $b_i$ of dimension $n$ with $n \ll k \times m$ where $k$
  corresponds to the encoding size of real-valued numbers (\textit{e.g.} 32 bits
  for standard single-precision floating-point format) while preserving the
  semantic information contained in $x_i$.  The proposed method achieves this
  objective by using an autoencoder model composed of two parts: an encoder
  that binarizes the word vector $x$ to $\Phi(x)$ and a decoder that
  reconstructs a real-valued vector from $\Phi(x)$.  The binarized vector is
  therefore the latent representation of the autoencoder. Figure \ref{architecture}
  summarizes this architecture.

  \subsubsection{Encoding to binary vectors}
  Let $W$ be a $n \times m$  matrix. We define the binarized vector $b_i$ of
  $x_i$ as:
  \begin{equation}
    b_i = \Phi(x_i) = h(W \cdot x_i^T)
  \end{equation}
  where $h(.)$ is an element-wise function that outputs a bit given a real value
  such as the Heaviside step function. The dimension of $W$ is $(n \times m)$
  i.e. the desired binary size and the size of the original real-valued vector.
  Therefore our model can be used to generate binary vectors of any size
  independently of the original size of the real-valued vector.

  \subsubsection{Reconstructing a real-valued vector}
  The generated representation $\Phi(x_i)$ is then used to compute a
  reconstructed vector $\hat{y}_i$ as:
  \begin{equation}
    \hat{y}_i = f(W^T \cdot \Phi(x_i) + c)
  \end{equation}
  where $c$ is a $m$-dimensional bias vector. The function $f$ is an
  element-wise non-linear function. The hyperbolic tangent function is used as
  the $f$ function in the model to be able to reconstruct the vectors as close
  as possible to the original embeddings $X$ whose values are within the [-1,1]
  range. Pre-trained vectors are clipped to be within the the [-1,1] range.
  This preprocess step does not alter the quality of the pre-trained vectors
  (semantic similarity score stays the same) as most of the vector values are
  already within this range.

\subsection{Objective function}
  The reconstruction loss $\ell_{rec}$ for a vector $x_i$ is defined as the mean
  squared error of $x_i - \hat{y}_i$:
  \begin{equation}
    \ell_{rec}(x_i) = {1 \over m} \sum_{k=0}^m (x_{i_k} - \hat{y}_{i_k})^2
  \end{equation}
  where $x_{i_k}$ (resp. $\hat{y}_{i_k}$) represents attribute number $k$ of
  vector $x_i$ (resp. $\hat{y}_{i}$). The autoencoder is trained to minimize
  this loss for all word vectors. We noticed that this model produced good
  reconstructed vectors but poor semantically related binary codes as it is
  solely trained to minimize the reconstruction loss. The learned matrix $W$ was
  discarding too much similarity information from the original space in favor of
  the reconstruction.

  This problem is solved by adding a regularization term in the objective
  function defined as:
  \begin{equation}
    \ell_{reg} = {1 \over 2} ||W^TW - I||^2
  \end{equation}
  This term tends to minimize the correlations between the different features of
  the latent binary representation which is an essential concept in our context.
  Since the model aims to produce small-size binary embeddings, it needs to
  encode as much information as possible across all the dimensions of the binary
  code, therefore minimizing the correlation between the different binary
  features is crucial to avoid duplicate information. This regularization has
  the effect of preserving the information contained in the real-valued
  embeddings, so vectors that are close in the original space are also close in
  the binary space.

  The balance between the influence of the reconstruction loss and the
  regularization loss is achieved with the $\lambda_{reg}$ parameter. The
  global objective function to minimize is:
  \begin{equation}
    \mathcal{L} = \sum_{x_i \in X} \ell_{rec}(x_i) + \lambda_{reg} \; \ell_{reg}
  \end{equation}
  Note here that the same parameter matrix $W$ is both used for the encoder and
  for the decoder. This is actually motivated by the fact that the function
  $h(\cdot)$ involved in the tranformation function $\Phi(\cdot)$ is
  non-differentiable, so it is not possible to compute a gradient for the encoding
  step. It is assumed that $\frac{\partial \Phi(x_i)}{\partial W} = 0$. However,
  the weights of $W$ used in the encoding step can still be updated thanks to
  the information provided by the decoder ($\frac{\partial (x_i -
  \hat{y}_i)^2}{\partial W}$). The autoencoder is then trained with stochastic
  gradient descent (SGD) with momentum set to 0.95.

  The objective function is both non-convex and non-smooth. Despite the fact
  that proving a general convergence is a hard problem, we have observed in
  practice that the regularization plays an important role and allows our model
  to converge to a local optimum (without the regularization, our model
  oscillates and binary vectors keep changing).

\section{Experimental setup}
  Several tasks have been run to measure the performance of the binary and
  reconstructed vectors (semantic similarity, word analogy, document
  classification, sentiment analysis, question classification) and another task
  to evaluate the computation efficiency of binary vectors (top-K queries).

  \subsubsection{Pre-trained embeddings}
  Our model produces binary vectors from several pre-trained embeddings:
  \texttt{dict2vec} \cite{tissier2017dict2vec} which contains 2.3M words and has
  been trained on the full English Wikipedia corpus; \texttt{fasttext}
  \cite{bojanowski2016enriching} which contains 1M words and has also been
  trained on the English Wikipedia corpus; and \texttt{GloVe}
  \cite{pennington2014glove} which contains 400k words and has been trained on
  both English Wikipedia and Gigaword 5 corpora. All vectors have 300
  dimensions. The learning hyperparameters used are the ones from their
  respective paper. Since the three embeddings are based on different methods
  (derivation of skip-gram for \texttt{dict2vec} and \texttt{fasttext}, matrix
  factorization for \texttt{GloVe}), this demonstrates that the model is general
  and works for all kinds of pre-trained vectors.

  \begin{table*}[t]
    \caption{Spearman's rank correlation similarity and word analogy scores
    for binary vectors ($bin$) of 64, 128, 256 and 512 bits, reconstructed
    real-valued vectors from the binary codes ($rec$) and binary vectors
    produced with Local Sensitive Hashing (LSH). For each dataset, scores
    of original real-valued vectors are also reported ($raw$ column).}
    \label{res:tasks-rank}
    \centering
    \resizebox{0.87\linewidth}{!}{
      \begin{tabular}{@{}lcccccrcccccrccccc@{}}

        \toprule[0.15em]
      & \multicolumn{5}{c}{dict2vec} & \phantom{}
      & \multicolumn{5}{c}{fasttext} & \phantom{}
      & \multicolumn{5}{c}{GloVe}\\

      \cmidrule{2-6} \cmidrule{8-12} \cmidrule{14-18}
      &$raw$&64&128&256&512&&$raw$&64&128&256&512&&$raw$&64&128&256&512\\
      \midrule
      MEN & 74.6 &&&&&& 80.7 &&&&&& 73.7\\
        \multicolumn{1}{r}{$bin$} &-& 66.1 & \textbf{71.3} & 70.3 & 71.3 &
                                  &-& 57.9 & 72.0 & 75.9 & \textbf{76.3} &
                                  &-& 46.1 & 63.3 & 69.4 & \textbf{72.7}\\
        \multicolumn{1}{r}{$rec$} &-& 64.5 & \underline{69.6} & 67.8 & 64.7 &
                                  &-& 45.8 & 56.2 & \underline{62.3} & 59.3 &
                                  &-& 43.6 & 50.5 & 68.5 & \underline{72.2}\\
        \multicolumn{1}{r}{LSH}   &-& 47.7 & 56.2 & 62.6 & 67.8 &
                                  &-& 47.7 & 60.7 & 70.0 & 75.0 &
                                  &-& 50.8 & 62.0 & 64.9 & 71.0\\
      RW & 50.5 &&&&&& 53.8 &&&&&& 41.2\\
        \multicolumn{1}{r}{$bin$} &-& 36.5 & 42.0 & 45.6 & 45.6 &
                                  &-& 36.8 & 44.7 & \textbf{52.7} & 52.7 &
                                  &-& 25.1 & 34.3 & \textbf{40.7} & 40.2\\
        \multicolumn{1}{r}{$rec$} &-& 35.7 & 41.6 & \underline{44.6} & 39.3 &
                                  &-& 28.9 & 31.6 & \underline{45.7} & 44.1 &
                                  &-& 24.8 & 29.2 & 36.4 & \underline{40.5}\\
        \multicolumn{1}{r}{LSH}   &-& 26.4 & 37.2 & 41.7 & \textbf{46.2} &
                                  &-& 34.5 & 40.3 & 47.5 & 46.5 &
                                  &-& 26.3 & 33.0 & 35.8 & 38.3\\
      SimLex & 45.2 &&&&&& 44.1 &&&&&& 37.1\\
        \multicolumn{1}{r}{$bin$} &-& 32.0 & 38.1 & \textbf{44.8} & 42.9 &
                                  &-& 25.1 & 38.0 & \textbf{44.6} & 43.0 &
                                  &-& 20.5 & 31.4 & \textbf{37.2} & 36.8\\
        \multicolumn{1}{r}{$rec$} &-& 30.4 & 37.9 & \underline{42.4} & 39.3 &
                                  &-& 19.2 & 30.0 & \underline{40.5} & 34.0 &
                                  &-& 19.6 & 19.1 & 34.2 & \underline{38.2}\\
        \multicolumn{1}{r}{LSH}   &-& 29.6 & 35.9 & 40.2 & 39.5 &
                                  &-& 28.7 & 32.0 & 38.6 & 41.1 &
                                  &-& 24.7 & 30.5 & 33.1 & 34.6\\
      SimVerb & 41.7 &&&&&& 35.6 &&&&&& 22.7\\
        \multicolumn{1}{r}{$bin$} &-& 25.3 & 36.6 & \textbf{38.4} & 35.5 &
                                  &-& 19.2 & 26.7 & 33.7 & \textbf{35.1} &
                                  &-&  7.8 & 18.7 & 22.9 & \textbf{23.0}\\
        \multicolumn{1}{r}{$rec$} &-& 23.7 & 35.4 & \underline{37.5} & 29.4 &
                                  &-& 12.8 & 18.2 & \underline{25.6} & 25.3 &
                                  &-&  8.0 & 12.4 & 22.1 & \underline{24.7}\\
        \multicolumn{1}{r}{LSH}   &-& 22.0 & 27.5 & 31.6 & 36.9 &
                                  &-& 20.5 & 23.3 & 30.7 & 30.2 &
                                  &-& 14.6 & 17.4 & 18.8 & 20.7\\
      WS353 & 72.5 &&&&&& 69.7 &&&&&& 60.9\\
        \multicolumn{1}{r}{$bin$} &-& 63.7 & \textbf{71.6} & 69.6 & 66.6 &
                                  &-& 50.3 & 69.1 & 70.0 & \textbf{70.3} &
                                  &-& 30.1 & 44.9 & 56.6 & \textbf{60.3}\\
        \multicolumn{1}{r}{$rec$} &-& 61.4 & \underline{69.0} & 67.4 & 58.8 &
                                  &-& 36.5 & 53.6 & \underline{64.0} & 53.6 &
                                  &-& 26.5 & 42.2 & 56.5 & \underline{62.0}\\
        \multicolumn{1}{r}{LSH}   &-& 45.5 & 56.9 & 64.9 & 65.5 &
                                  &-& 46.7 & 53.2 & 58.6 & 63.8 &
                                  &-& 41.1 & 44.4 & 50.5 & 57.8\\
      \midrule
      Sem. analogy & 59.6 &&&&&& 37.6 &&&&&& 77.4\\
        \multicolumn{1}{r}{$bin$} &-& 2.6 & 12.0 & 26.7 & \textbf{30.1} &
                                  &-& 2.3 & 7.5  & 18.0 & \textbf{25.0} &
                                  &-& 8.5 & 26.7 & 53.4 & \textbf{65.3}\\
        \multicolumn{1}{r}{$rec$} &-& 2.6 & 10.2 & 22.8 & \underline{30.9} &
                                  &-& 1.8 &  5.0 & 14.6 & \underline{15.2} &
                                  &-& 7.7 & 23.0 & 49.1 & \underline{62.8}\\
        \multicolumn{1}{r}{LSH}   &-& 0.8 &  4.6 & 14.9 & 29.9 &
                                  &-& 0.8 &  6.4 & 13.0 & 20.4 &
                                  &-& 6.1 & 25.9 & 47.9 & 59.3\\
      Syn. analogy & 54.0 &&&&&& 87.4 &&&&&& 67.0\\
        \multicolumn{1}{r}{$bin$} &-& 3.5 & 16.7 & 34.8 & \textbf{36.2} &
                                  &-& 8.0 & 34.5 & 57.3 & 64.7 &
                                  &-& 7.3 & 23.9 & 46.3 & \textbf{52.4}\\
        \multicolumn{1}{r}{$rec$} &-& 3.6 & 16.1 & 31.2 & \underline{37.5} &
                                  &-& 4.6 & 14.6 & 50.8 & \underline{53.1} &
                                  &-& 7.3 & 21.7 & 44.6 & \underline{54.0}\\
        \multicolumn{1}{r}{LSH}   &-& 1.7 &  7.8 & 23.4 & 35.8 &
                                  &-& 4.0 & 21.5 & 46.2 & \textbf{65.7} &
                                  &-& 5.6 & 21.6 & 39.1 & 52.3\\
      \bottomrule[0.15em]
    \end{tabular}}
  \end{table*}

  \subsubsection{Training settings}
  Binary vectors of 4 different sizes are produced: 64, 128, 256 and 512 bits.
  The optimal hyperparameters are found using a grid search and selected to
  minimize the reconstruction loss and the regularization loss described in
  Section \ref{binauto}. The model uses a batch size of 75, 10 epochs for
  dict2vec and fasttext, and 5 epochs for GloVe (the autoencoder converges
  faster due to the smaller vocabulary) and a learning rate of 0.001.  The
  regularization hyperparameter $\lambda_{reg}$ depends on the starting vectors
  and the binary vector size. It varies from 1 to 4 in the experiments but its
  influence on the performance is small ($\sim$2\% variation).

  \subsubsection{Semantic word similarity}
  \label{protocol_sim}
  Both binary and reconstructed vectors are evaluated with the standard method
  which consists in computing the Spearman's rank correlation coefficient
  between the similarity scores attributed by humans to pairs of words and the
  scores computed with the word vectors of these pairs. The score of real-valued
  vectors is the cosine similarity and the score of binary vectors is the Sokal
  \& Michener similarity function \cite{sokal1958statistical} defined as:
  \begin{equation}
    sim(v_1, v_2) = {n_{11} + n_{00} \over {n}}
  \end{equation}
  where $n_{11}$ (resp. $n_{00}$) is the number of bits in $v_1$ and $v_2$ that
  are both set to 1 (resp. set to 0) simultaneously and $n$ is the vector size.
  The similarity datasets used are MEN \cite{bruni2014multimodal}, RW
  \cite{luong2013better}, SimLex \cite{hill2015simlex}, SimVerb
  \cite{gerz2016simverb} and WordSim \cite{finkelstein2001placing}.

  \subsubsection{Word analogy}
  This evaluation follows the standard protocol used by Mikolov et al.
  \shortcite{mikolov2013distributed}. The task consists in finding the word $d$
  in questions like ``$a$ is to $b$ as $c$ is to $d$''. The evaluation first
  computes the vector $v_b - v_a + v_c$ and then look at its closest
  neighbours.  If the closest one is the vector associated to $d$, then the
  analogy is correct. The task reports the fraction of correctly guessed
  analogies among all analogies. For binary vectors, we replace the addition
  with the \texttt{OR} bitwise operator and the subtraction with the \texttt{AND
  NOT} operator because adding or subtracting bits do not really make sense in
  the binary space (like subtracting the bit 1 to the bit 0). The dataset is
  separated in 2: one consists of analogies about countries and currencies
  (semantic), the other one about english grammar (syntactic).

  \begin{table*}[t]
    \caption{Document classification (top), question classification (middle) and
    sentiment analysis (bottom) accuracies for binary vectors ($bin$) of 64,
    128, 256 and 512 bits, reconstructed real-valued vectors ($rec$) and binary
    vectors produced with Local Sensitive Hashing (LSH).
    For each dataset, scores of original real-valued  vectors are also reported
    ($raw$ column).}
    \label{res:tasks-classif}
    \centering
    \resizebox{0.87\linewidth}{!}{
      \begin{tabular}{@{}lcccccrcccccrccccc@{}}
        \toprule[0.15em]
      & \multicolumn{5}{c}{dict2vec} & \phantom{}
      & \multicolumn{5}{c}{fasttext} & \phantom{}
      & \multicolumn{5}{c}{GloVe}\\
      \cmidrule{2-6} \cmidrule{8-12} \cmidrule{14-18}
      &$raw$&64&128&256&512&&$raw$&64&128&256&512&&$raw$&64&128&256&512\\
      \midrule
      AG-News & 89.0 &&&&&& 86.9 &&&&&& 89.5\\
        \multicolumn{1}{r}{$bin$} &-& 85.3 & 85.9 & 87.7 & 87.8 &
                                  &-& 84.5 & 85.9 & 87.3 & 87.7 &
                                  &-& 84.0 & 87.2 & 88.5 & 88.5\\
        \multicolumn{1}{r}{$rec$} &-& 85.2 & 86.3 & \underline{87.9} & 87.2 &
                                  &-& 82.8 & 84.3 & \underline{87.7} & 87.3 &
                                  &-& 83.9 & 87.7 & 88.6 & \underline{89.2}\\
        \multicolumn{1}{r}{LSH}   &-& 78.8 & 82.6 & 86.1 & \textbf{88.1} &
                                  &-& 77.5 & 83.3 & 86.1 & \textbf{88.8} &
                                  &-& 83.5 & 86.6 & 88.4 & \textbf{88.6}\\
      DBpedia & 97.6 &&&&&& 95.0 &&&&&& 97.2\\
        \multicolumn{1}{r}{$bin$} &-& 94.1 & 96.1 & 97.0 & 97.3 &
                                  &-& 91.7 & 95.1 & 96.6 & \textbf{97.3} &
                                  &-& 90.9 & 95.0 & 96.8 & \textbf{97.2}\\
        \multicolumn{1}{r}{$rec$} &-& 94.0 & 95.9 & \underline{96.8} & 96.6 &
                                  &-& 89.5 & 92.6 & \underline{96.4} & 96.0 &
                                  &-& 91.2 & 95.2 & 96.8 & \underline{97.0}\\
        \multicolumn{1}{r}{LSH}   &-& 89.6 & 94.2 & 96.5 & \textbf{97.4} &
                                  &-& 87.4 & 93.8 & 96.2 & 97.2 &
                                  &-& 90.4 & 94.2 & 96.3 & 97.2\\
      \midrule
      Yahoo Ans & 68.1 &&&&&& 67.2 &&&&&& 68.1\\
        \multicolumn{1}{r}{$bin$} &-& 60.7 & 63.8 & 66.0 & 66.8 &
                                  &-& 60.4 & 63.9 & 66.4 & \textbf{67.8} &
                                  &-& 57.5 & 62.5 & 66.4 & 66.1\\
        \multicolumn{1}{r}{$rec$} &-& 60.8 & 63.8 & 65.9 & \underline{66.0} &
                                  &-& 60.0 & 62.9 & 66.3 & \underline{66.8} &
                                  &-& 58.4 & 64.3 & 66.7 & \underline{67.0}\\
        \multicolumn{1}{r}{LSH}   &-& 52.3 & 59.9 & 64.5 & \textbf{67.1} &
                                  &-& 52.2 & 59.5 & 64.9 & 66.9 &
                                  &-& 56.8 & 62.0 & 65.3 & \textbf{67.0}\\
      \midrule
      Amazon Full & 47.5 &&&&&& 49.0 &&&&&& 47.1\\
        \multicolumn{1}{r}{$bin$} &-& 39.9 & 43.9 & 46.8 & 47.7 &
                                  &-& 39.0 & 43.9 & 47.9 & \textbf{49.8} &
                                  &-& 37.4 & 42.6 & 46.7 & 47.8\\
        \multicolumn{1}{r}{$rec$} &-& 40.1 & 44.0 & \underline{46.8} & 46.2 &
                                  &-& 39.1 & 43.8 & 48.1 & \underline{48.5} &
                                  &-& 39.8 & 45.3 & 47.1 & \underline{47.3}\\
        \multicolumn{1}{r}{LSH}   &-& 38.3 & 42.5 & 45.6 & \textbf{48.1} &
                                  &-& 38.6 & 42.7 & 47.3 & 49.5 &
                                  &-& 37.9 & 43.0 & 45.9 & \textbf{48.6}\\
      Amazon Pol & 84.2 &&&&&& 85.6 &&&&&& 83.8\\
        \multicolumn{1}{r}{$bin$} &-& 76.3 & 80.7 & 83.2 & 83.8 &
                                  &-& 75.1 & 80.2 & 84.5 & \textbf{85.8} &
                                  &-& 73.1 & 78.9 & 83.2 & 84.4\\
        \multicolumn{1}{r}{$rec$} &-& 76.6 & 80.8 & \underline{83.2} & 82.4 &
                                  &-& 75.1 & 80.2 & 84.7 & \underline{84.8} &
                                  &-& 76.6 & 80.2 & 83.6 & \underline{83.7}\\
        \multicolumn{1}{r}{LSH}   &-& 74.3 & 79.0 & 81.9 & \textbf{84.5} &
                                  &-& 73.8 & 78.5 & 83.4 & 85.7 &
                                  &-& 74.7 & 79.1 & 82.1 & \textbf{85.0}\\
      Yelp Full & 52.5 &&&&&& 52.1 &&&&&& 52.7\\
        \multicolumn{1}{r}{$bin$} &-& 45.1 & 48.7 & 51.6 & 52.0 &
                                  &-& 44.2 & 49.7 & 53.0 & \textbf{54.6} &
                                  &-& 42.7 & 48.4 & 51.8 & 53.2\\
        \multicolumn{1}{r}{$rec$} &-& 45.3 & 48.8 & \underline{51.6} & 50.9 &
                                  &-& 43.5 & 47.8 & 53.0 & \underline{53.1} &
                                  &-& 43.4 & 50.3 & 52.3 & \underline{52.8}\\
        \multicolumn{1}{r}{LSH}   &-& 43.0 & 47.7 & 51.0 & \textbf{53.1} &
                                  &-& 44.3 & 47.6 & 52.4 & 54.3 &
                                  &-& 43.6 & 48.2 & 51.5 & \textbf{53.4}\\
      Yelp Pol & 87.8 &&&&&& 88.0 &&&&&& 87.9\\
        \multicolumn{1}{r}{$bin$} &-& 80.8 & 84.5 & 86.6 & 87.6 &
                                  &-& 80.1 & 84.5 & 88.1 & 89.5 &
                                  &-& 77.8 & 84.2 & 86.9 & \textbf{88.7}\\
        \multicolumn{1}{r}{$rec$} &-& 80.9 & 84.5 & \underline{86.6} & 86.1 &
                                  &-& 79.6 & 84.0 & 88.2 & \underline{88.5} &
                                  &-& 78.6 & 85.7 & 87.5 & \underline{87.7}\\
        \multicolumn{1}{r}{LSH}   &-& 77.9 & 82.8 & 86.1 & \textbf{88.0} &
                                  &-& 80.3 & 82.2 & 87.2 & \textbf{89.8} &
                                  &-& 79.0 & 83.1 & 86.6 & 88.6\\

      \bottomrule[0.15em]
    \end{tabular}}
  \end{table*}

  \begin{figure*}[t] 
    \begin{center}
      \centerline{\includegraphics[width=\linewidth]{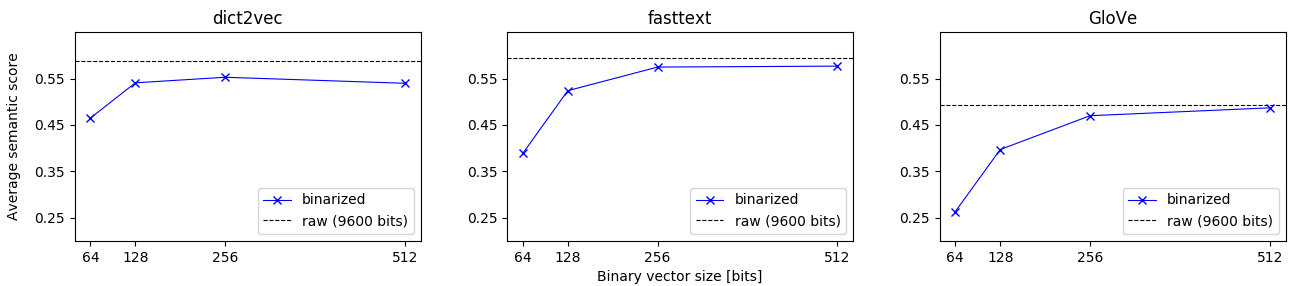}}
      \caption{Fisher's transformed average semantic correlation scores for
      different binary vector size for all type of embeddings. The \emph{raw}
      baseline indicates the score obtained with real-valued embeddings.}
      \label{scores_plot}
    \end{center}
  \end{figure*}

  \subsubsection{Document/Question classification and sentiment analysis}
  The evaluation follows the same protocol as described in the literature
  \cite{zhang2015character,joulin2016bag} which consists in
  predicting the assigned label given the bag-of-words representation of a text.
  A single hidden layer neural network where the input weights have been
  initialized with the binary or reconstructed vectors is used. The input
  weights are fixed during training so that the classification accuracy only
  depends on the vectors used to initialize the neural network. The datasets
  used are AG-News and DBpedia for document classication, Yahoo Answers for
  question classification and Amazon and Yelp reviews (both polarity and full)
  for the sentiment analysis task. Each dataset is
  split into a training and a test file and the same training and test files are
  used for all word embedding models. Accuracy results are reported in Table
  \ref{res:tasks-classif}.

  \subsubsection{Top-K queries performances}
  A top-K query consists in finding the \emph{K} closest vectors given a single
  word vector query. The closest vectors are the ones with the
  highest similarity with the query vector (Sokal \& Michener similarity for
  binary vectors, cosine similarity for real-valued ones). The top-K vectors are
  found by performing a linear scan across all vectors. Two execution times are
  measured for both binary and real-valued vectors: the time it takes to get
  the results once the vectors are loaded in memory and the time it takes to
  load the vectors and perform the query.

\section{Results and binary vectors analysis}
  \label{results}

\subsection{Binary embeddings performances}
  \subsubsection{Semantic word similarity}
  Table \ref{res:tasks-rank} reports the Spearman's rank correlation scores
  obtained with the binarized vectors (\emph{bin}) and the scores of the
  original real-valued vectors (\emph{raw}) whose size is 9600 bits (300
  dimensions, 32-bit floating-point values). The best scores for binarized
  vectors are reached with 256 or 512 bits. For fasttext and GloVe, the results
  are very close to the scores obtained with the raw vectors (absolute
  difference smaller than 1 point). For dict2vec, the deviation is larger
  (between 3 and 5 points) but still in the same order of magnitude.

  Binarizing word vectors can lead to better scores compared to the original
  real-valued ones (the \emph{raw} column). For instance, the 512-bit version of
  fasttext on WS353 (70.3 against 69.7) or the 256-bit version of GloVe on
  SimVerb (22.9 against 22.7). Moreover, the 64-bit version of dict2vect is
  better than the real-valued GloVe embedding (which uses 9600 bits of
  information per vector, so the binary codes are 150 times smaller) on SimVerb
  (25.3 against 22.7) and WS353 (63.7 against 60.9). This demonstrates that the
  method can produce rich semantically related small binary codes but more
  generally, that \emph{binary vectors can have the same semantic information as
  the real-valued vectors}.

  The average semantic correlation scores (aggregated with the Fisher's
  transformation) of binary and original vectors (whose size is 9600 bits per
  vector) are plotted in Figure \ref{scores_plot}. The 512-bit version of GloVe is
  on par with the real-valued version. However the performance loss is greater
  for fasttext and dict2vec vectors. These embeddings contain 1M of words for
  fasttext, and 2.3M for dict2vec, whereas GloVe only contains 400k words. As
  the binary space has a finite size depending on the binary vector size, it
  becomes harder for the autoencoder to find a distribution of the binary
  vectors that preserves the semantic similarity when the number of vectors
  increases.

  \subsubsection{Word Analogy}
  Table \ref{res:tasks-rank} reports the word analogy accuracies of binary
  vectors for both semantic and syntactic analogies. Although the best scores
  are also obtained with larger binary codes (512 bits), the scores are lower
  than the real-valued vectors. This can be explained by the fact that the task
  of word analogy is not suited to evaluate binary vectors. With real-valued
  vectors, the analogies are found with vector addition and subtraction in the
  $\mathcal{R}^d$ space. In the binary space (where each value is either 0 or
  1), adding or subtracting vectors does not make sense (like subtracting the
  bit 1 from the bit 0) resulting in poor accuracies for the word analogies
  with the binary vectors.

  \subsubsection{Text classification}
  The accuracies obtained on the different datasets of the document
  classification task for both binarized vectors (\emph{bin}) and original
  vectors (\emph{raw}) are provided in Table \ref{res:tasks-classif}. For this
  task, the binarized vectors achieve the best scores with 512 bits in general.
  Similarly to the semantic similarity task, the binary vectors are sometimes
  better than the original vectors. This is especially true for the fasttext
  binarized vectors where the accuracies goes from 95.0 to 97.3 on DBpedia, or
  goes from 49.0 to 49.8 on Amazon Full reviews.

  Smaller sizes of binary vectors lead to better compression rates but cause a
  slight decrease in performance. The 256-bit vectors of GloVe are 37.5 times
  smaller than the original vectors (whose size is 9600 bits) but have an
  accuracy drop between 0.4\% and 2.5\% depending on the dataset. The 64-bit
  vectors of dict2vec (compression rate of 150) have a loss of accuracy of about 4\%
  on AG-News and DBpedia, about 11\% on Yahoo answers and about 16\% on Amazon
  Full reviews.  The two latter datasets are the bigger ones with 3M training samples
  for Amazon Full reviews and 1.4M for Yahoo answers while AG-News and DBpedia
  respectively contain 120k and 560k training samples. As the dataset becomes
  larger, the information required to correctly classify the documents also
  increases, so is the accuracy loss on these datasets when small binary vectors
  are used.  Overall, this experiments show once again that our binary embeddings are
  competitive in comparison with real-valued ones.

\subsection{Reconstructed vectors performances}
  Table \ref{res:tasks-rank} and Table \ref{res:tasks-classif} also respectively
  report the scores obtained on semantic word similarity and analogy and text
  classification tasks using the reconstructed ($rec$) vectors.

  On the semantic similarity task, fasttext and dict2vec mostly best operate
  using 256-bit binary vectors while GloVe requires larger binary vectors
  (512-bit) to reconstruct good real-valued vectors. For most datasets, the
  reconstructed vectors from the binarized GloVe 512-bit representations
  outperforms the original real-valued vectors: 38.2 against 37.1 on SimLex,
  24.7 against 22.7 on SimVerb and 62.0 against 60.9 on WS353. For dict2vec and
  fasttext, binarizing and then reconstructing real-valued vectors from the
  binary ones causes a loss of performance compared to the original word vectors:
  between -4.8\% (WS353) and -11.7\% (RW) for dict2vec; between -8.2\% (WS353)
  and -28.1\% (SimVerb) for fasttext. The performance loss of reconstructed
  vectors is larger for fasttext than for dict2vec due to their different
  embedding scheme. Fasttext also encodes additional morphological information
  in the word vector space by considering that a word is the sum of its
  subwords, which is harder to reconstruct after a dimension reduction (the
  binarization).

  On NLP application tasks like document classification or sentiment
  analysis (Table \ref{res:tasks-classif}), the results are very consistent:
  the reconstructed vectors exhibit close performances to the binary
  representations, which in turn exhibit (almost) equal performances to the
  original vectors -- whatever the initial embedding model. Although the
  pairwise word semantic similarity is not perfectly preserved in reconstructed
  vectors, the document classification task only needs vectors close enough
  for those of the same category but not necessarily very accurate within a
  category. This makes the binary or reconstructed vectors good for a real use
  case NLP task. The optimal number of bits required to recontruct good
  real-valued vector is the same as for the semantic similarity task (256 bits
  for dict2vec/fasttext, 512 for GloVe).

\subsection{Speed improvements in top-K queries}
  The execution time (in milliseconds) of top-K queries benchmarks for GloVe
  vectors are reported in Table \ref{topk_benchmark}. The first three rows (Top
  1, Top 10, Top 50) indicate the time used by the program to perform the query
  \emph{after} the vectors are loaded \emph{i.e.} the time to linearly scan all
  the vectors, compute the similarity with the query vector and select the K
  highest values.  Finding the closest 10 words with 256-bit vectors is 30 times
  faster compared to using real-valued vectors and can be up to 36 times faster
  with the 64-bit vectors.

  Having faster computations is not the only interest of binary vectors. Since
  they take less space in memory, they are also much faster to load. The last
  line in Table \ref{topk_benchmark} indicates the time needed to load the
  vectors from a file and perform a top-10 query. It takes 310 milliseconds to
  load the 256-bit binary vectors and run the query whereas it take 23.5 seconds
  for the real-valued ones, which is 75 times slower.

  \begin{table}[h]
    \caption{Execution time (in ms) to run a top-K query on binary and real-valued
    vectors.}
    \label{topk_benchmark}
    \centering
    \resizebox{\linewidth}{!}{
    \begin{tabular}{cccccc}
      \toprule
      Execution time (ms) & 64-bit & 128-bit & 256-bit & 512-bit & Real-valued\\
      \midrule
      Top 1               &  2.71  &   2.87  &   3.23  &   4.28  &   97.89    \\
      Top 10              &  2.72  &   2.89  &   3.25  &   4.29  &   98.08    \\
      Top 50              &  2.75  &   2.91  &   3.27  &   4.32  &   98.44    \\
      Loading + Top 10 & 160  & 213  &   310   &   500   &   23500    \\
      \bottomrule
    \end{tabular}}
  \end{table}

\subsection{Comparison with the naive approach}
  \label{ssection:naive}
  A naive approach to produce binary vectors is to map negative values of
  pre-trained word embeddings to 0 and positive values to 1. Unfortunately the
  binary vectors produced with this method have the same number of dimensions as
  the original vectors and since most pre-trained vectors are not available in
  64, 128 or 256 dimensions, the binary vectors are not aligned with CPU
  registers (see Section \ref{intro}) which is an important requirement to achieve
  fast vector operations.

  To produce aligned vectors with the naive method, one has first to train
  real-valued embeddings with 64, 128 or 256 dimensions and then apply a naive
  binarization function to all vector values. First, this process is slower than
  our method. It requires 8 hours while we only need 13 minutes. Second, binary
  vectors produced with the naive binarization perform worse compared to those
  produced with the proposed method. The Fisher's average semantic score is 44.7
  and 49.6 for the naive 64 and 128 bits vectors while the 64 and 128 bits
  dict2vec vectors achieve 46.5 and 54.1.

\subsection{Comparison with other binarization methods}
  We used the Locality Sensitive Hashing method to produce binary vectors
  of 64, 128, 256 and 512 bits from the same real-valued vectors as us and have
  evaluated them on the same tasks (Table 1 and 2). For the word similarity and
  analogy tasks, our vectors are always better (e.g. 70.3 vs. 63.8 for
  fasttext 512-bit on WS353, 34.8 vs. 23.4 for dict2vec 256-bit on Syn.
  analogy) except for GloVe 64-bit and some fasttext 64-bit vectors. For the
  classification task, our 512-bit vectors are on par with the LSH ones, but our
  128 and 256 bits vectors have better performances than the respective LSH
  ones. Our model is better suited when the compression is higher and gives
  the best size to performance ratio.

  We also used the Faruqui method \shortcite{faruqui2015sparse} to produce
  binary vectors. Faruqui's vectors are mostly sparse (90\% of bits set to 0)
  but not very small to be computationally interesting: dimension is set to
  3000-bit, so they do not fit in registers. They are also have poorer
  performance than our binary vectors: 0.560 for their average score on semantic
  similarity while we achieve 0.575 with our 256-bit vectors.

\subsection{Qualitative analysis of binary embeddings}
  \subsubsection{Evolution of word vector similarity} The semantic word
  similarity task rely on specific datasets. They contain pairs of words and a
  value assigned by humans (\emph{e.g.} computer/keyboard - 0.762)). This value
  represents the semantic similarity between the two words and is assumed to be
  the ground truth that word embeddings should encode. The cosine similarity of
  the pre-trained vectors of the words of one pair does not always match the
  value associated to this pair, but binarizing the vectors helps to move closer
  the vector similarity to the value, making the embedding closer to the human
  judgment. Some pairs are reported in Table \ref{outliers}.
  \begin{table}[h]
    \caption{Semantic similarity for some pairs of words, evaluated by human or
    computed with binary/real-valued vectors.}
    \label{outliers}
    \centering
    \resizebox{0.9\linewidth}{!}{
    \begin{tabular}{cccc}
      \toprule
      Words              & Human judgment & Binary & Cosine  \\
      \midrule
      dollar -- buck     &    0.92        &  0.56  &  0.13   \\
      seafood -- sea     &    0.75        &  0.62  &  0.24   \\
      money -- salary    &    0.79        &  0.58  &  0.41   \\
      car -- automobile  &    0.98        &  0.69  &  0.60   \\
      \bottomrule
    \end{tabular}}
  \end{table}

  \begin{figure}[t]
    \begin{center}
      \centerline{\includegraphics[width=\linewidth]{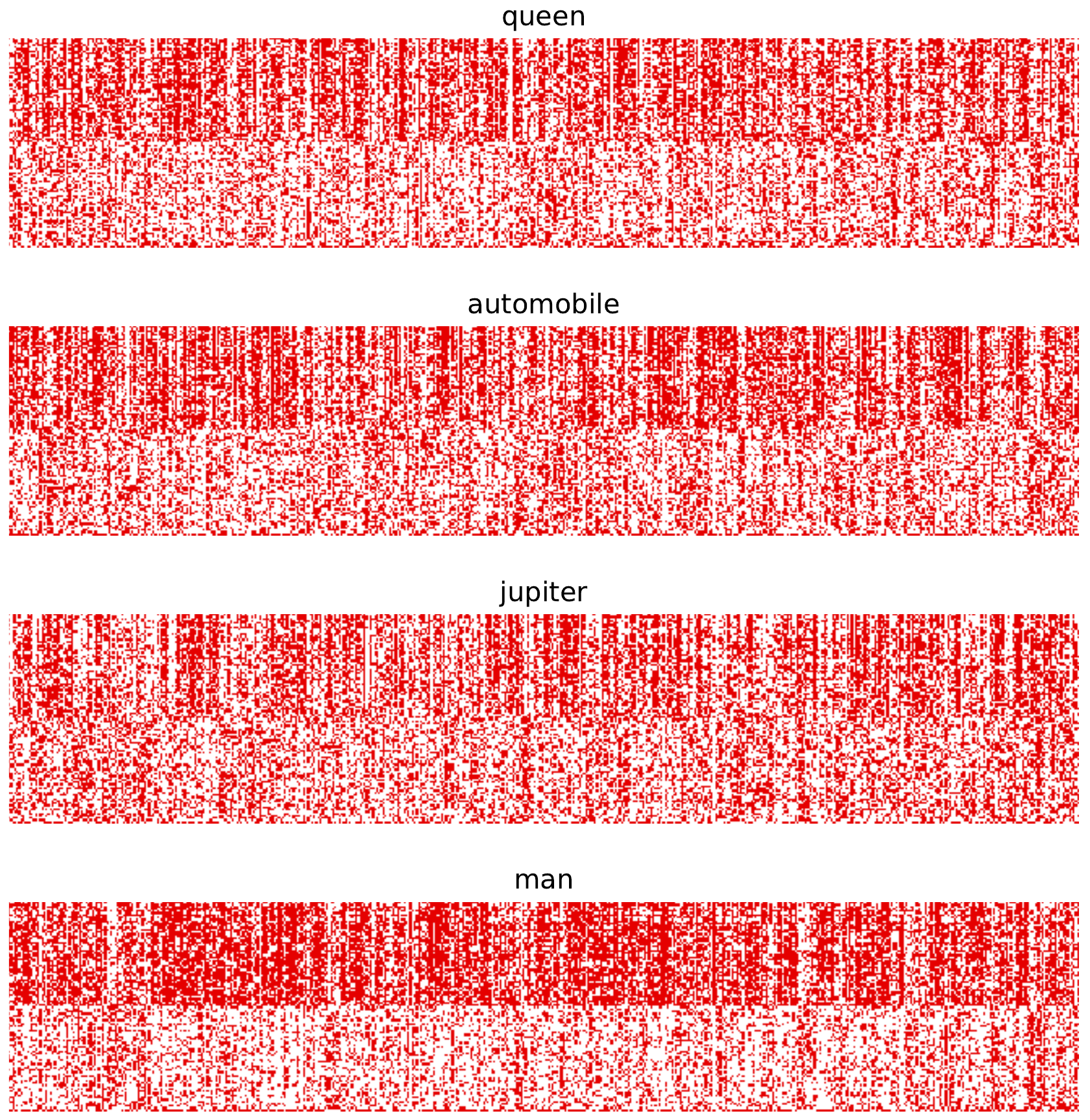}}
      \caption{Representation of the activated bits in some binary vectors for
      the words queen, automobile, jupiter and man. For
      each word, its 50 closest neighbors (top) and its 50 furthest neighbors
      (bottom) are plotted. Red pixels represent the bits set to 1 and white
      pixels the bits set to 0.}
      \label{visualization}
    \end{center}
  \end{figure}

  \subsubsection{Visualization of activated dimensions in binary vectors} In
  Figure \ref{visualization}, the 50 most similar (top) and the 50 least similar
  (bottom) 512-bit binary vectors of some words are plotted. Each pixel is
  either red or white depending on whether it is 1 or 0. Some vertical
  stripes are clearly visible in the top area: similar word vectors have the
  same bits enabled for certain dimensions and therefore share the same color
  of pixel for those dimensions. This is especially true for the closest words
  of ``automobile''.  In the bottom area, no vertical patterns are visible
  because two non-similar binary vectors do not have the same enabled bits.
  This visualization shows that the binarized vectors have semantic properties
  similar to real-valued embeddings.

\section{Conclusion}
  \label{conclusion}
  This article presents a new encoder/decoder architecture for transforming
  real-valued vectors into binary vectors of any size. This is particularly
  suitable when the size of the binary vectors corresponds to the CPU cache
  sizes as it allows a significant increase in vector operation computing speed.
  Our method has the advantage of being simple yet powerful and allows to keep
  semantic information in binary vectors.

  Our binary embeddings exhibit almost the same performances as the original
  real-valued vectors on both semantic similarity and document classification
  tasks. Furthermore, since the binary representations are smaller (a 256-bit
  binary code is 37.5 smaller than a traditional 300-dimensional word vector),
  it allows one to run a top-K query 30 times faster than with real-valued
  vectors since the binary distance is much faster to compute than a regular
  cosine distance. Additionally, it is possible to recontruct real-valued
  vectors from the binary representations using the model decoder.

\bibliography{aaai19}
\bibliographystyle{aaai19}
\end{document}